\documentclass{svmult}

\usepackage{mathptmx}
\usepackage{helvet}
\usepackage{courier}
\usepackage{makeidx}
\usepackage{amsmath,amsfonts}
\usepackage{graphicx}
\usepackage{cite}
\usepackage{centercaption}
\usepackage[bottom]{footmisc}

\usepackage{xmpmulti}
\usepackage{multirow}
\usepackage{amsbsy}	

\makeatletter
\newcommand\notsotiny{\@setfontsize\notsotiny\@vipt\@viipt}
\makeatother

\begin{document}

\title*{A Comparative Study on 2-DOF Variable Stiffness Mechanisms}
\author{Christoph Stoeffler \and Shivesh Kumar \and Andreas M\"uller}
\institute{%
 Christoph Stoeffler \at
 Universit\"at Bremen - AG Robotik,
 \email{stoeffle@uni-bremen.de}
  \\
 Shivesh Kumar \at
 DFKI Bremen - Robotics Innovation center,
 \email{shivesh.kumar@dfki.de}
 \\
 Andreas M\"uller \at
 Johannes Kepler University, Linz, Austria,
 \email{a.mueller@jku.at}
}

\authorrunning{Stoeffler et al.}

\maketitle

\vspace{-2.5cm}
\abstract{%
Based on the idea of \textit{variable stiffness mechanisms}, a variety of such mechanisms is shown in this work. Specifically, 2-DOF \textit{parallel kinematic machines} equipped with redundant actuators and non-linear springs in the actuated joints are presented and a comparative overview is drawn. Accordingly, a general stiffness formulation in task space of all mechanisms is given. Under fixed geometric parameters, optimization of task space stiffness is carried out on the designs comprising all kinematic solutions. Finally, a stiffness metric is introduced that allows a quantitative comparison of the given mechanism designs. This gives rise to design guidelines for engineers but also shows an interesting outline for future applications of variable stiffness mechanisms.} 


\vspace{-0.7cm}
\section{Introduction and overview}
\label{sec:introduction}
\vspace{-0.4cm}

 Using \textit{parallel kinematic machines} (PKMs) in modern robotic designs results in many advantages, due to the possibility of redundant actuation, higher load capacity and so on. Another recent tendency in robotics is the integration of flexible and soft links and joints to comply with higher safety standards, but also to increase the performance of robotic systems - may it be through energy storage or reduced mass, where for the latter flexibility is rather a consequence than an enhancement. Most notably, \textit{serial elastic actuators} (SEAs) and \textit{variable impedance actuators} (VIAs) find many applications in nowadays systems - see e.g~\cite{ham_compliant_2009}. Combining the conceptual ideas of PKMs and VIAs leads to a mechanism type that we call \textit{variable stiffness mechanisms} (VSMs) and which we recently proposed in \cite{stoeffler_conceptual_2018}. In general, VSMs allow simultaneous position and stiffness control of several end-effector DOFs and can thus be seen as a higher dimensional extension of VIAs. It is hoped that such designs lead to further enhancements of robotic systems, as they integrate compliance into multi-DOF mechanisms and give rise to impact tolerance, energy storage and stiffness adaptations. In this work a comparative overview of 2-degree-of-freedom (2-DOF) VSMs based on stiffness bounds is presented. For this purpose various possible VSMs are presented in this section followed by a generic approach of \textit{Cartesian stiffness} modeling of all presented VSMs in Section~\ref{sec:models}. A metric to quantify the stiffness in task space is presented and computed for all mechanisms in Section~\ref{sec:simulations} that is used to derive some design guidelines. Finally, Section~\ref{sec:conclusion} draws conclusions and future perspectives of this work.\\

\begin{figure}
  \centering
  \includegraphics[width=0.85\textwidth]{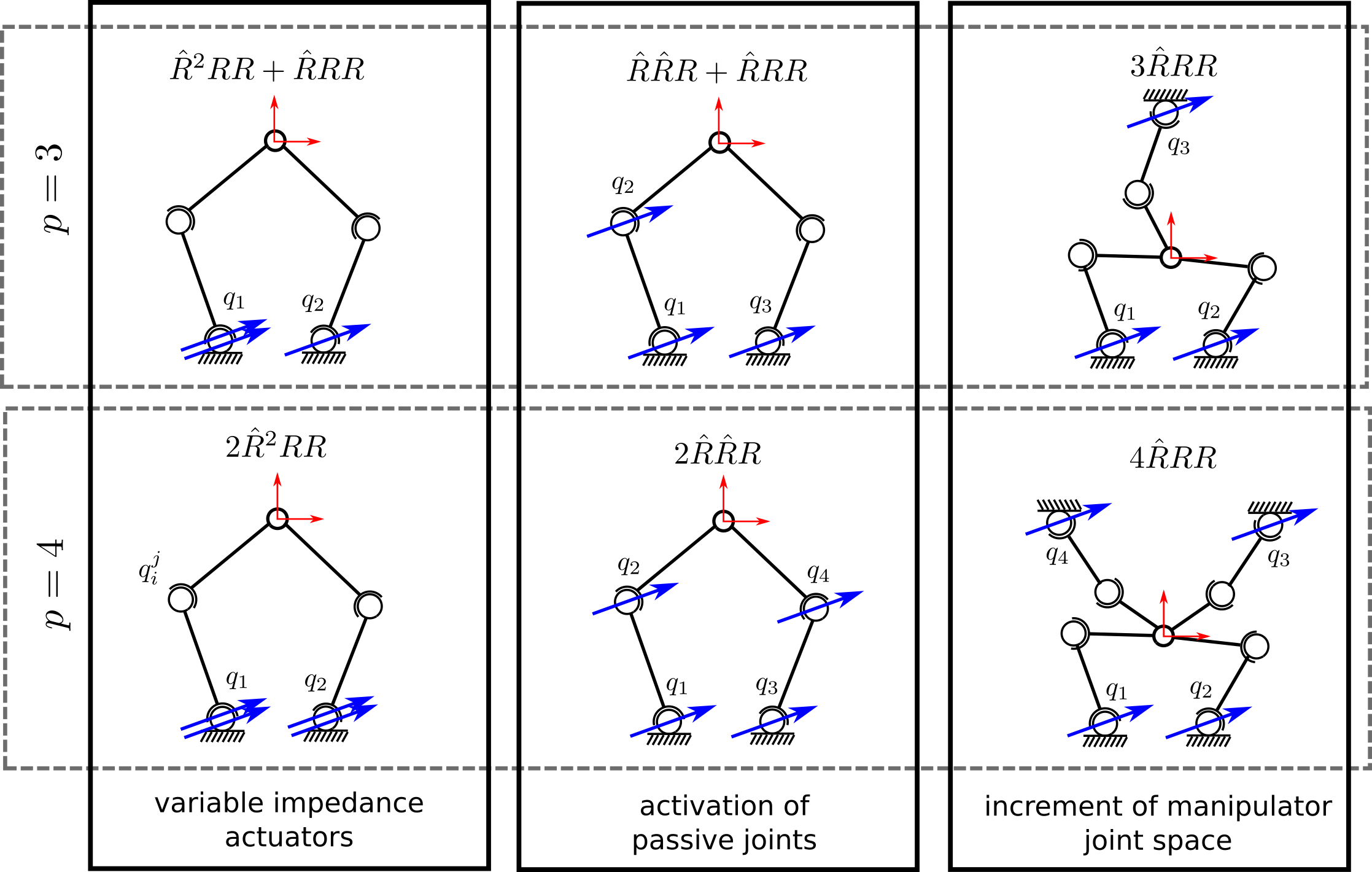}
  \caption{Overview of six different planar 2-DOF VSM, distinguished by the degree of actuation $p$ and the way it is achieved. Arrows indicate installed motor-spring combinations - single arrow: (nonlinear) SEA, double arrow: VIA. A mechanism can be described by full coordinates $q_i^j$ ($j$th joint coordinate in $i$th leg) or by $m$ generalized coordinates $q_k$ respectively.}
  \label{fig:VSM_overview_6_flat}
\end{figure}

Like animal muscle groups, VSMs rely on antagonistic actuation within parallel structures that can be achieved in different ways. An early example of this approach can be found in~\cite{yi_open-loop_1989}, where stiffness control is opted via antagonistic actuation and later e.g. in~\cite{muller_stiffness_2006}, including the dynamics of the system. Figure~\ref{fig:VSM_overview_6_flat} shows six different ways~\footnote{For a 2-DOF mechanism, this is the number of reasonable designs, as we omit designs that are symmetric to others and that have higher degree of redundancy than end-effector DOFs.} to achieve a planar 2-DOF VSM, ordered by the degree of actuation and its implementation.
Arrows in Figure~\ref{fig:VSM_overview_6_flat} indicate a module of non-linear spring combined with a drive. Pairs of those in one joint indicate pair of springs and drives what would typically be a VIA. For the sake of stiffness changes at the end-effector, non-linear springs are necessary - stiffness changes could be induced by the non-linearity and antagonism of the mechanism itself, like e.g. in~\cite{kock_parallel_1998, muller_stiffness_2006}, but we demand here, that the passive stiffness can also be changed. This in turn requires stiffness changes on joint level, what is expected to also enlarge the possible stiffnesses in the task space. A description of different redundancies in PKMs is given in~\cite{obrien_redundant_1999} with the terms \textit{activation of passive joints} and \textit{increment of manipulator joint space} that apply to the centre box and right box of Figure~\ref{fig:VSM_overview_6_flat} respectively. But we can also exploit redundant actuation with VIAs in the base joints - shown in the left box of that figure.

\vspace{-0.7cm}
\section{Stiffness modeling and optimization}
\label{sec:models}
\vspace{-0.4cm}
Stiffness is the coefficient in the first order approximation of the relation of force and displacement. It is thus determined by the first-order term of the \textit{Taylor expansion} of a force $\vec{f}$ with respect to a displacement $\vec{s}$ 
whereas, usually higher-order terms are neglected and then stiffness generally writes
\begin{equation}
 \vec{K}_s = \frac{d\vec{f}}{d\vec{s}}. \label{equ:stiffness}
\end{equation}
We can define stiffness in different spaces, but it is required that $\vec{f}$ and $d\vec{s}$ exist in a certain space and its \textit{dual}. This naturally results in square stiffness matrices, where off-diagonal terms describe the coupling between the coordinates contained in $\vec{s}$. When force changes are decoupled from each other, diagonal stiffness matrices result. Those are e.g. present when the joint space stiffness of robotic systems is considered. A widely discussed problem is the conversion of stiffness from joint space to task space e.g. $f: \vec{K}_{q} \mapsto \vec{K}_x$. A good overview that addresses this mapping is given in~\cite{pashkevich_enhanced_2011} encompassing different works on serial and parallel manipulators. One major difference in the accounted works is the treatment of external loading that is commonly said to be responsible for asymmetric stiffness matrices. It has been argued by Hoevenaars and others~\cite{hoevenaars_consistent_2016} that asymmetry is a modeling inconsistency, as the resulting matrices have to arise from \textit{conservative} properties alone. Zefran and Kumar~\cite{zefran_geometric_2002} show in a geometric setting that asymmetry is resolved by choosing an \textit{affine connection} on the robots configuration - the Lie group $SE(3)$. In contrast to this, Chen~\cite{chen_spatial_2005} illustrates asymmetric matrices as a result of \textit{non-coordinate transformations}. What follows, is a practical argumentation for a model of symmetric stiffness matrices of our designs, exemplified by the \textit{conservative congruence transformation} (CCT)~\cite{chen_simulation_2000}. Mapping via Jacobian between joint space forces $\vec{\tau}$ and task space forces $\vec{f}$ is generally
\begin{equation}
 \pmb{\tau} = \vec{J}^T\vec{f} \label{equ:force}
\end{equation}
Differentiating and making use of the stiffness expression (\ref{equ:stiffness}) yields
\begin{eqnarray}
 d\pmb{\tau} = \left(d\vec{J}^T\right)\vec{f}+\vec{J}^T\left(d\vec{f}\right) \label{equ:diff_force} \\[5pt]
 \vec{K}_{q}d\vec{q} = \left(\frac{\partial\vec{J}^T}{\partial\vec{q}}d\vec{q}\right)\vec{f}+\vec{J}^T\vec{K}_xd\vec{x} \nonumber 
\end{eqnarray}
where the first term on the right hand side can be changed by reordering indices and the second via expressing $d\vec{x}$ by the Jacobian
\begin{equation}
 \vec{K}_{q}d\vec{q} = \underbrace{\left[\left(\frac{\partial\vec{J}^T}{\partial q_1}\vec{f}\right)\quad\left(\frac{\partial\vec{J}^T}{\partial q_2}\vec{f}\right)\quad ... \quad \left(\frac{\partial\vec{J}^T}{\partial q_m}\vec{f}\right)\right]}_\text{\normalsize $\vec{K}_g$}d\vec{q}+\vec{J}^T\vec{K}_x\vec{J}d\vec{q} \nonumber 
\end{equation}
Finally, rearranging gives the task space stiffness
\begin{equation}
  \vec{K}_x = \vec{J}^{-T}(\vec{K}_{q}-\vec{K}_g)\vec{J}^{-1} \label{equ:cct}
\end{equation}
where $\vec{K}_{q}$ is the diagonal joint space stiffness matrix and $\vec{K}_g$ captures the Jacobian changes from external loading. 

\begin{remark}
Whether $\vec{K}_g$ should be included or not in (\ref{equ:cct}) depends on the assumption whether a certain configuration of the mechanism can be maintained or not. In our proposed designs, spring deflections can be compensated by the serially attached motors. Thus external forces lead to stiffness changes on joint level, but the configuration of the mechanism is depending on commanded position $\vec{x}_c$ alone and $\left(d\vec{J}^T\right)\vec{f}$ in (\ref{equ:diff_force}) vanishes. The stiffness mapping in (\ref{equ:cct}) simplifies to
\begin{equation}
 \vec{K}_x = \vec{J}^{-T}\vec{K}_{q}\vec{J}^{-1} \label{equ:cct_simple}
\end{equation}
External loading can thus be seen as a problem that is solely reflected in $\vec{K}_{q}$ what is in accordance with~\text{\cite{hoevenaars_consistent_2016}}.
\end{remark}

\vspace{-0.7cm}
\subsection{Constraint Jacobian}
\label{sec:jacobian}
\vspace{-0.4cm}
The Jacobian matrix of a mechanism is given by differentiation of a constraint equation $\vec{g}(\vec{x},\vec{q})=\vec{0}$ such as
\begin{eqnarray}
 \frac{\partial\vec{g}(\vec{x},\vec{q})}{\partial\vec{x}}\vec{\dot{x}} = 
 -\frac{\partial\vec{g}(\vec{x},\vec{q})}{\partial\vec{q}}\vec{\dot{q}} \\[5pt]
 \vec{\dot{x}} = -\vec{J}_x^{\dagger}\vec{J}_{q}\vec{\dot{q}} \label{equ:invert}\\[5pt]
 \vec{\dot{x}} = \vec{J}\vec{\dot{q}}
\end{eqnarray}
where $\vec{J}_x^{\dagger}$ is the \textit{pseudoinverse} in the case of $dim(\vec{g})>dim(\vec{x})$  and otherwise the usual inverse of $\vec{J}_x$.

\vspace{-0.7cm}
\subsection{Stiffness computation in task space}
\label{sec:stiffness_computation}
\vspace{-0.4cm}
With the above equations and $i$ legs attached to the end-effector, task space stiffness can be computed according to the scheme in Figure~2, where the functions $f_i^{-1}: \vec{x}\mapsto q_i^j$ and $\partial f_i^{-1}/\partial\vec{x}: q_i^j\mapsto\vec{J}_i^{leg}$ are the inverse and differential kinematics of 2-DOF serial planar mechanisms, as to be found e.g. in~\cite{lynch_modern_2017}. It should be noted that the full state of the mechanism $q_i^j$ is computed and the subset $q_k$ is used to obtain the Jacobians of the parallel mechanism $\vec{J}$ to account for external loading. Finally, Cartesian stiffness is obtained by the sum of individual leg stiffnesses, where passive joints are accounted by zero joint stiffness.
\begin{figure}
 \centering
 \includegraphics[width=0.9\textwidth]{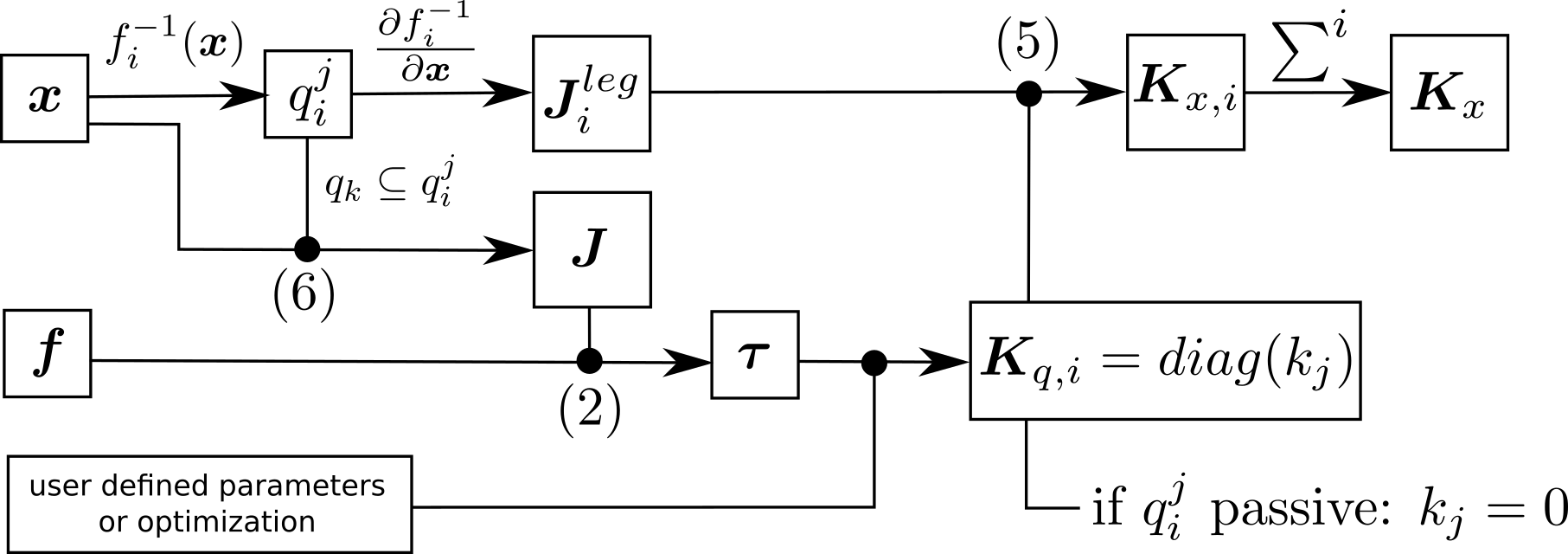}
 \label{fig:scheme}
 \caption{Depiction of stiffness computation in task space with optimized or user defined stiffness. Connections denote the application of related functions to compute the boxed entities.}
\end{figure}

\vspace{-0.7cm}
\subsection{Jacobian null space and joint space stiffness optimization}
\label{sec:nullspace}
\vspace{-0.4cm}

Redundancy is reflected in the Jacobian matrix of (\ref{equ:invert}) and can be directly used for stiffness optimization in task space. A general solution of (\ref{equ:force}) is presented through
\begin{equation}
 \pmb{\tau} = \vec{J}^T\vec{f}+\left(\vec{I}_n-\vec{J}^T\vec{J}^{\dagger T}\right)\pmb{\tau}_n \label{equ:operational_space_formulation}
\end{equation}
with $\vec{I}$ as the identity matrix and $\pmb{\tau}_n$ being an arbitrary vector in the space of $\pmb{\tau}$. Since the second term of (\ref{equ:operational_space_formulation}) represents the null space of $\vec{J}$ and thus $\pmb{\tau}_n$ has no impact on $\vec{f}$, one could also write
\begin{equation}
 \pmb{\tau} = \vec{J}^T\vec{f}+\ker(\vec{J})\vec{u} \label{equ:kernel}
\end{equation}
where $\ker(\vec{J})$ gives the basis of the null space of $\vec{J}$ and $\vec{u}$ is an arbitrary vector with dimension equal to the degree of redundancy. Optimizing Cartesian stiffness by spring deflection $\gamma_i$ for a prescribed configuration then writes
\begin{table}[h]
  \centering
 \normalsize
\vspace{-0.4cm}
 \begin{tabular}{c c l}
  \textbf{maximize} & \hspace{0.4cm} & $\det(\vec{K}_x(\vec{x}, \gamma_1, \dots, \gamma_m))$ 	\\
  \small{$\gamma_1, \dots, \gamma_m$, $\vec{u}$} 				& & \\[8pt]
  \textbf{subject to} & & $\vec{\tau}(\gamma_1, \dots, \gamma_m) = \vec{J}^T\vec{f}+\ker(\vec{J})\vec{u}$ \\[5pt]
  & & $-\gamma_{max} < \gamma_i < \gamma_{max}$
 \end{tabular}
\end{table}

\vspace{-0.5cm}
\noindent In designs using VIAs (left box in Figure~\ref{fig:VSM_overview_6_flat}), the objective function reduces to simple joint stiffness $k_{j}=d\tau^{s1}/d\gamma_1+d\tau^{s2}/d\gamma_2$ where $\tau^{s1}$ and $\tau^{s2}$ are the spring forces in the VIA and the constraints reduce to a simple force equilibrium $\tau_i-\tau^{s1}+\tau^{s2}=0$. The optimized joint space stiffness $\vec{K}_{q}$ is then mapped to task space as in the scheme above.

\vspace{-0.7cm}
\section{A metric for work space stiffness}
\label{sec:metric}
\vspace{-0.4cm}

By looking at the determinant of Cartesian stiffness matrices, a scalar value implying magnitude of the underlying transformation can be given. Taking the difference between maximized and minimized stiffness further informs about the achievable difference, since absolute magnitude can always be influenced by installing different springs in the joints. As the variations in magnitude can be of some orders of magnitude, it is further appropriate to take the logarithm, thus \textit{stiffness variation} writes
\begin{eqnarray}
 sv = \log(\det(\vec{K}_x)^{max})-\log(\det(\vec{K}_x)^{min}) \nonumber \\[5pt]
 =\log\left(\frac{\det(\vec{K}_x)^{max}}{\det(\vec{K}_x)^{min}}\right)
\end{eqnarray}
By integrating the variation over work space and normalizing, a configuration dependent \textit{work space stiffness metric} is obtained
\begin{equation}
 wssm = \frac{\int\,sv(\vec{x})dV}{\int\,dV}
\end{equation}

\vspace{-0.7cm}
\section{Simulations and discussion}
\label{sec:simulations}
\vspace{-0.4cm}

\begin{figure}
  \centering
  \includegraphics[width=0.65\textwidth]{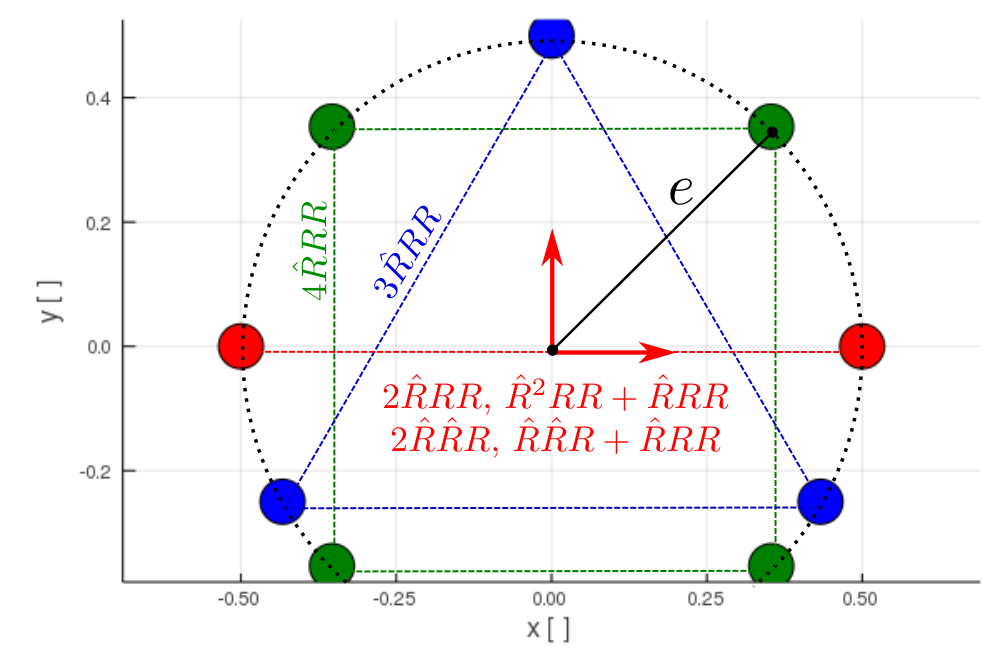}
  \caption{Base joints (dots) lie on a circle with radius $e$ in symmetric arrangements. Geometric parameters are normalized to the circles diameter. The coordinate system shows the work space origin.}
  \label{fig:base_geometry}
\end{figure}

\begin{figure}
  \centering
  \includegraphics[width=1.0\textwidth]{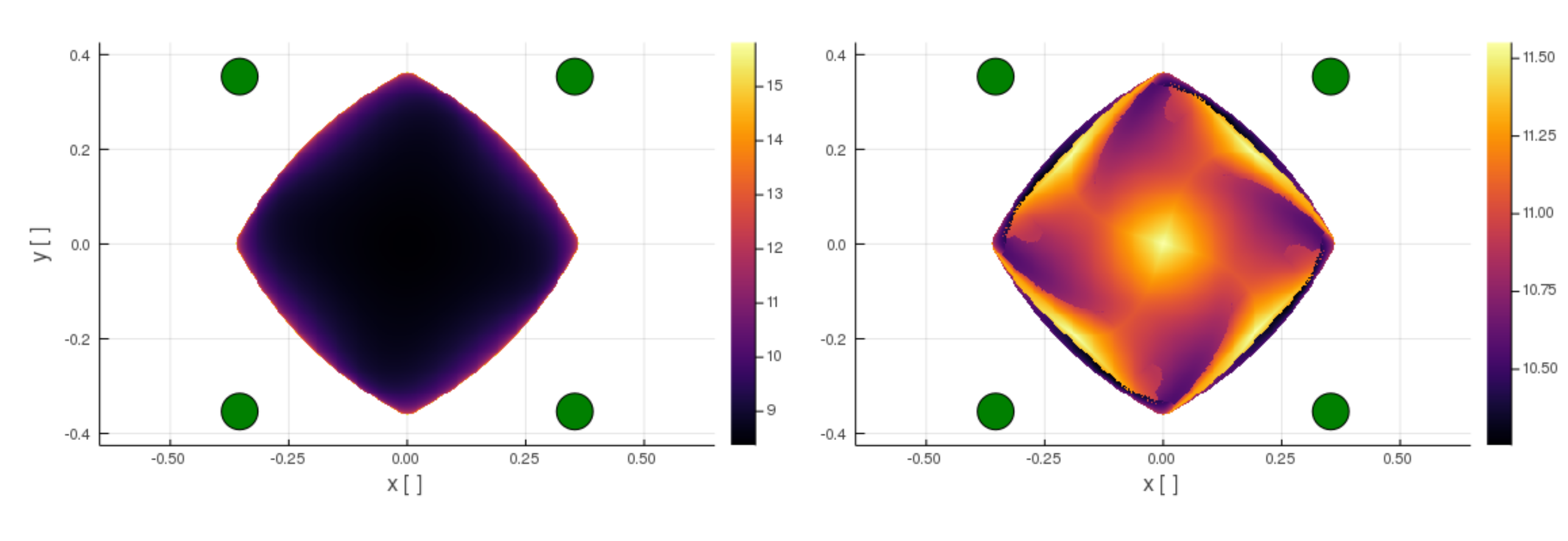}
  \caption{Minimized (left) and maximized (right) $\log(\det(\vec{K}_x))$ of $4\hat{R}RR:r=0.4$, $l=0.4$, $e=0.5$ of the first kinematic solution (out of 16)}
  \label{fig:optimal_stiffness_4Rrr_2}
\end{figure}

To allow a comparison between all designs, location of the base joints is fixed as depicted in Figure~\ref{fig:base_geometry}, leaving $r$, $l$ and $e$ as geometric parameters that are kept dimensionless for simulations.
Springs are modeled by means of the point-symmetric force function $\tau^s=5\sin(\gamma)\cdot\gamma^3$ in the interval $\gamma\in[-\pi, \pi]$ that results in a convex stiffness function $d\tau^s/d\gamma$ throughout this region. Even though, our formulations account for external forces, stiffness bounds of the different designs are computed without loading. In most cases and without external forces, stiffness minimization can simply be computed with $\vec{K}_{q}^{min}=f(\gamma_i=0)$ and mapped to Cartesian space. Points in task space with a condition index $1/\text{cond}(\vec{J}\vec{J}^T)<10^{-4}$ are discarded from computation to avoid optimization attempts close to singularities. In Figure~\ref{fig:optimal_stiffness_4Rrr_2} a stiffness optimization of $4\hat{R}RR$ is shown. 
The values of the minimized solution increase at the boundaries, since one of the legs gets into a stretched position, possessing high stiffness in one direction. A similar phenomenon can be seen with the $3\hat{R}RR$. Table~\ref{tab:metric} contains all computed metrics of the investigated mechanisms for two arbitrary geometric settings. It can be observed, that the computed metric of designs involving VIAs ($2\hat{R}^2RR$ and $\hat{R}^2RR+\hat{R}RR$)  in the base is independent of geometry and kinematic solution - the magnitude is solely reflected by the non-linear function of the springs. Moreover, the metric of mechanism $2\hat{R}^2RR$ with two VIAs, has twice the magnitude as that of $\hat{R}^2RR+\hat{R}RR$ which accounts that two eigenvalues of $\vec{K}_x$ can be changed independently instead of one.  
\begin{table}
\notsotiny
\centering
 \caption{Work space stiffness metric for all designs and kinematic solutions. Numbers in the vectors indicate first or second inverse solution of the independent joint coordinates.}
 \begin{tabular}{l|c|c|c|c|c|c|c|c|c|c|c|c|c|c|c|c|c}
 & \multicolumn{17}{c}{\textbf{kinematic solutions}} \\
 \textbf{mechanism}
 & $\begin{bmatrix} 1 \\ 1 \\ 1 \\ 1 \end{bmatrix}$ 
 & $\begin{bmatrix} 2 \\ 1 \\ 1 \\ 1 \end{bmatrix}$
 & $\begin{bmatrix} 1 \\ 2 \\ 1 \\ 1 \end{bmatrix}$
 & $\begin{bmatrix} 2 \\ 2 \\ 1 \\ 1 \end{bmatrix}$
 & $\begin{bmatrix} 1 \\ 1 \\ 2 \\ 1 \end{bmatrix}$
 & $\begin{bmatrix} 2 \\ 1 \\ 2 \\ 1 \end{bmatrix}$
 & $\begin{bmatrix} 1 \\ 2 \\ 2 \\ 1 \end{bmatrix}$
 & $\begin{bmatrix} 2 \\ 2 \\ 2 \\ 1 \end{bmatrix}$
 & $\begin{bmatrix} 1 \\ 1 \\ 1 \\ 2 \end{bmatrix}$
 & $\begin{bmatrix} 2 \\ 1 \\ 1 \\ 2 \end{bmatrix}$
 & $\begin{bmatrix} 1 \\ 2 \\ 1 \\ 2 \end{bmatrix}$
 & $\begin{bmatrix} 2 \\ 2 \\ 1 \\ 2 \end{bmatrix}$
 & $\begin{bmatrix} 1 \\ 1 \\ 2 \\ 2 \end{bmatrix}$
 & $\begin{bmatrix} 2 \\ 1 \\ 2 \\ 2 \end{bmatrix}$
 & $\begin{bmatrix} 1 \\ 2 \\ 2 \\ 2 \end{bmatrix}$
 & $\begin{bmatrix} 2 \\ 2 \\ 2 \\ 2 \end{bmatrix}$
 & $\begin{bmatrix} q_1 \\ q_2 \\ q_3 \\ q_4 \end{bmatrix}$  \\[10pt]
 \hline
 \multicolumn{18}{c}{\textbf{geometric parameters:} $r_i=r=0.5$, $l_i=l=0.25$, $e=0.5$} \\[5pt]
 $\hat{R}^2RR+\hat{R}RR$         & 1.59 & 1.59 & 1.59 & 1.59 & - & - & - & - & - & - & - & - & - & - & - & - & \\
 $2\hat{R}^2RR$                  & 3.19 & 3.19 & 3.19 & 3.19 & - & - & - & - & - & - & - & - & - & - & - & - & \\
 $\hat{R}\hat{R}R+\hat{R}RR$     & 2.04 & 2.02 & 2.02 & 2.04 & - & - & - & - & - & - & - & - & - & - & - & - & \\
 $2\hat{R}\hat{R}R$              & 2.61 & 2.60 & 2.60 & 2.61 & - & - & - & - & - & - & - & - & - & - & - & - & \\
 $3\hat{R}RR$                    & 2.22 & 2.35 & 2.35 & 2.35 & 2.35 & 2.35 & 2.35 & 2.22 & - & - & - & - & - & - & - & - & \\
 $4\hat{R}RR$                    & 2.63 & 2.54 & 2.54 & 2.63 & 2.54 & 2.62 & 2.63 & 2.54 & 2.54 & 2.63 & 2.62 & 2.54 & 2.63 & 2.54 & 2.54 & 2.63 & \\
 \hline
 \multicolumn{18}{c}{\textbf{geometric parameters:} $r_i=r=0.4$, $l_i=l=0.4$, $e=0.5$} \\[5pt]
 $\hat{R}^2RR+\hat{R}RR$         & 1.59 & 1.59 & 1.59 & 1.59 & - & - & - & - & - & - & - & - & - & - & - & - & \\
 $2\hat{R}^2RR$                  & 3.19 & 3.19 & 3.19 & 3.19 & - & - & - & - & - & - & - & - & - & - & - & - & \\
 $\hat{R}\hat{R}R+\hat{R}RR$     & 2.25 & 2.05 & 2.05 & 2.25 & - & - & - & - & - & - & - & - & - & - & - & - & \\
 $2\hat{R}\hat{R}R$              & 2.79 & 2.79 & 2.79 & 2.79 & - & - & - & - & - & - & - & - & - & - & - & - & \\
 $3\hat{R}RR$                    & 2.35 & 2.13 & 2.13 & 2.13 & 2.13 & 2.13 & 2.13 & 2.35 & - & - & - & - & - & - & - & - & \\
 $4\hat{R}RR$                    & 2.61 & 2.43 & 2.43 & 2.70 & 2.43 & 2.65 & 2.70 & 2.43 & 2.43 & 2.70 & 2.65 & 2.43 & 2.70 & 2.43 & 2.43 & 2.61 &
 \end{tabular}
 \label{tab:metric}
\end{table}
Under the present geometries and in terms of stiffness variations, it is preferable to locate actuators in the base when the degree of redundancy is of order two ($2\hat{R}^2RR$ compared to $2\hat{R}\hat{R}R$ and $4\hat{R}RR$), while $4\hat{R}RR$ allows the removal of singularities in a then more confined workspace. This is in contrast to simple redundancy, where $\hat{R}^2RR+\hat{R}RR$  possesses smaller stiffness variations than $\hat{R}\hat{R}R+\hat{R}RR$ and $3\hat{R}RR$. Because of the activation of passive joints, $2\hat{R}\hat{R}R$ and $\hat{R}\hat{R}R+\hat{R}RR$ usually have mass distributions that are not to the advantage of dynamic movements. In terms of stiffness changes, it can however be of interest to use the design $\hat{R}\hat{R}R+\hat{R}RR$, since it allows bigger stiffness variations than $\hat{R}^2RR+\hat{R}RR$ but maintaining the same work space size.

\vspace{-0.7cm}
\section{Conclusion}
\label{sec:conclusion}
\vspace{-0.4cm}

In this work, a variety of planar 2-DOF VSMs has been introduced. The modeling applies to general PKM. The optimization of Cartesian stiffness in task space has been carried out. To this end, a metric has been introduced to assess and compare the different designs. We believe that the properties of such mechanisms should be further investigated, as their advantages can only be exploited with a thorough theoretical understanding.

\begin{acknowledgement}
Funded by the Deutsche Forschungsgemeinschaft (DFG, German Research Foundation) - 404971005. The third author acknowledges the support from the “LCM – K2 Center for Symbiotic Mechatronics” within the framework of the Austrian COMET-K2 program.
\end{acknowledgement}

\bibliographystyle{spmpsci}
\vspace{-0.8cm}
\bibliography{bibfile}

\end{document}